\documentclass[11pt,a4paper]{article}
\usepackage[hyperref]{emnlp-ijcnlp-2019}
\usepackage[T1,T5]{fontenc}
\usepackage[utf8]{inputenc}
\usepackage{tipa}
\usepackage{amsmath}
\usepackage{amssymb}
\usepackage{array}
\usepackage{booktabs}
\usepackage{multirow}
\usepackage{times}
\usepackage{latexsym}
\usepackage{graphicx}
\usepackage{float}
\usepackage{tikz-dependency}
\usepackage{subcaption}
\usepackage{placeins}
\usepackage{ragged2e}
\usepackage{paralist}
\usepackage[colorinlistoftodos]{todonotes}
\usepackage{url}
\usepackage{cleveref}
\usepackage{lingmacros}
\usepackage{anyfontsize}

\newcommand{\mytilde}{\raise.17ex\hbox{$\scriptstyle\mathtt{\sim}$}}

\newcommand*{\affaddr}[1]{#1} 
\newcommand*{\affmark}[1][*]{\textsuperscript{#1}}
\newcommand*{\email}[1]{\texttt{#1}}

\crefformat{section}{\S#2#1#3} 
\crefformat{subsection}{\S#2#1#3}
\crefformat{subsubsection}{\S#2#1#3}

\aclfinalcopy 

\title{A systematic comparison of methods for low-resource dependency parsing\\ on genuinely low-resource languages}

\author{%
Clara Vania\affmark[1] \quad
Yova Kementchedjhieva\affmark[2] \quad
Anders Søgaard\affmark[2] \quad
Adam Lopez\affmark[1] \\
\affaddr{\affmark[1]School of Informatics, University of Edinburgh, UK} \\
\affaddr{\affmark[2]University of Copenhagen, Copenhagen, Denmark} \\
\email{\small c.vania@ed.ac.uk, \{yova|soegaard\}@di.ku.dk, alopez@inf.ed.ac.uk} \\
}

\date{}
\begin{document}
\maketitle

\begin{abstract}

Parsers are available for only a handful of the world's languages, since they require lots of training data. How far can we get with just a small amount of training data? 
We systematically compare a set of simple strategies for improving low-resource parsers: data augmentation, which has not been tested before; cross-lingual training; and transliteration. Experimenting on three typologically diverse low-resource languages---North Sámi, Galician, and Kazah---We find that (1) when only the low-resource treebank is available, data augmentation is very helpful; (2) when a related high-resource treebank is available, cross-lingual training is helpful and complements data augmentation; and (3) when the high-resource treebank uses a different writing system, transliteration into a shared orthographic spaces is also very helpful.

\end{abstract}

\section{Introduction}

Large annotated treebanks are available for only a tiny fraction of the world’s languages, and there is a wealth of literature on strategies for parsing with few resources \citep{Hwa:2005:BPV:1088141.1088144,zeman2008,mcdonald-emnlp11,P11-2120}. A popular approach is to train a parser on a related high-resource language and adapt it to the low-resource language. This approach benefits from the availability of Universal Dependencies \citep[UD;][]{NIVRE16.348}, prompting substantial research \citep{Tiedemann:2016:STC:3013558.3013565,agic17,K18-2019}, along with the VarDial and the CoNLL UD shared tasks \citep{vardial,K17-3001,K18-2001}.

But low-resource parsing is still difficult. The organizers of the CoNLL 2018 UD shared task \citep{K18-2001} report that, in general, results on the task's nine low-resource treebanks ``are extremely low and the outputs are hardly useful for downstream applications.’’ So if we want to build a parser in a language with few resources, what can we do?
To answer this question, we systematically compare several practical strategies for low-resource parsing, asking:
\begin{enumerate}
    \item What can we do with only a very small \textit{target} treebank for a low-resource language?
    \item What can we do if we also have a \textit{source} treebank for a related high-resource language?
    \item What if the source and target treebanks do not share a writing system?
\end{enumerate}
Each of these scenarios requires different approaches. \textbf{Data augmentation} is applicable in all scenarios, and has proven useful for low-resource NLP in general \citep{augmentation-mt:ACL2017,bergmanis-etAl:K17-2002,sahin-emnlp18}. Transfer learning via \textbf{cross-lingual training} is applicable in scenarios 2 and 3. Finally, \textbf{transliteration} may be useful in scenario 3.

To keep our scenarios as realistic as possible, we assume that no taggers are available since this would entail substantial annotation. Therefore, our neural parsing models must learn to parse from words or characters---that is, they must be \textbf{lexicalized}---even though there may be little shared vocabulary between source and target treebanks. While this may intuitively seem to make cross-lingual training difficult, recent results have shown that lexical parameter sharing on characters and words can in fact improve cross-lingual parsing \citep{deLhoneux-emnlp18}; and that in some circumstances, a lexicalized parser can outperform a delexicalized one, even in a low-resource setting \citep{W17-6303}. 

We experiment on three language pairs from different language families, in which the first of each is a genuinely low-resource language: North Sámi and Finnish (Uralic); Galician and Portuguese (Romance); and Kazakh and Turkish (Turkic), which have different writing systems\footnote{We select high-resource language based on language family, since it is the most straightforward way to define language relatedness. However, other measurement (e.g., WALS \citep{wals} properties) might be used.}. To avoid optimistic evaluation, we extensively experiment only with North Sámi, which we also analyse to understand \emph{why} our cross-lingual training outperforms the other parsing strategies. We treat Galician and Kazakh as truly held-out, and test only our best methods on these languages. Our results show that:
\begin{enumerate}
    \item When no source treebank is available, data augmentation is very helpful: dependency tree morphing improves labeled attachment score (LAS) by as much as 9.3\%. Our analysis suggests that syntactic rather than lexical variation is most useful for data augmentation. 
    \item When a source treebank is available, cross-lingual parsing improves LAS up to 16.2\%, but data augmentation still helps, by an additional 2.6\%. Our analysis suggests that improvements from cross-lingual parsing occur because the parser learns syntactic regularities about word order, since it does not have access to POS and has little reusable information about word forms.
    \item If source and target treebanks have different writing systems, transliterating them to a common orthography is very effective. 
\end{enumerate}

\section{Methods}
\label{sec:data-aug}

We describe three techniques for improving low-resource parsing: (1) two data augmentation methods which have not been applied before for dependency parsing, (2) cross-lingual training, and (3) transliteration. 

\subsection{Data augmentation by dependency tree morphing (Morph)} 

\begin{figure}[t]
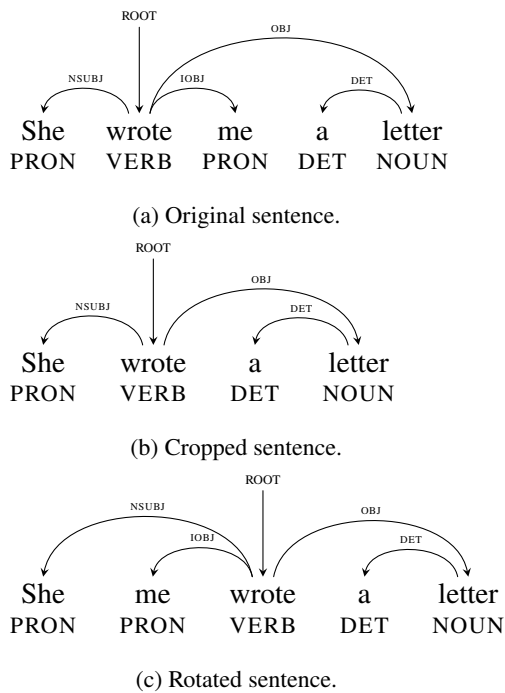

\centering
    \captionsetup[subfigure]{justification=centering}
    \begin{subfigure}[b]{0.4\textwidth}
    \begin{dependency}[theme=simple]
        \begin{deptext}[column sep=0.2cm]
            She \& wrote \& me \& a \& letter \\
            \textsc{pron} \& \textsc{verb} \& \textsc{pron} \& \textsc{det} \& \textsc{noun} \\
        \end{deptext}
    \deproot[edge unit distance=2ex]{2}{ROOT}
    \depedge{2}{1}{\textsc{nsubj}}
    \depedge{2}{3}{\textsc{iobj}}
    \depedge{5}{4}{\textsc{det}}
    \depedge{2}{5}{\textsc{obj}}
    \end{dependency}
    \caption{Original sentence.}
    \label{fig:tm-orig-sent}
    \end{subfigure}
    
    \begin{subfigure}[b]{0.4\textwidth}
    \begin{dependency}[theme = simple]
        \begin{deptext}[column sep=1em]
          She \& wrote \& a \& letter \\
          \textsc{pron} \& \textsc{verb} \& \textsc{det} \& \textsc{noun} \\
        \end{deptext}
    \deproot[edge unit distance=2ex]{2}{ROOT}
    \depedge{2}{1}{\textsc{nsubj}}
    \depedge{2}{4}{\textsc{obj}}
    \depedge{4}{3}{\textsc{det}}
    \end{dependency}
    \caption{Cropped sentence.}
    \label{fig:tm-cropped-sent}
    \end{subfigure}
    
    \begin{subfigure}[b]{0.4\textwidth}
    \begin{dependency}[theme = simple]
        \begin{deptext}[column sep=1em]
          She \& me \& wrote \& a \& letter \\
          \textsc{pron} \& \textsc{pron} \& \textsc{verb} \& \textsc{det} \& \textsc{noun} \\
        \end{deptext}
    \deproot[edge unit distance=2ex]{3}{ROOT}
    \depedge[edge height=3cm]{3}{2}{\textsc{iobj}}
    \depedge{3}{1}{\textsc{nsubj}}
    \depedge{3}{5}{\textsc{obj}}
    \depedge{5}{4}{\textsc{det}}
    \end{dependency}
    \caption{Rotated sentence.}
    \label{fig:tm-rotated-sent}
    \end{subfigure}
\caption{Examples of dependency tree morphing operations on the sentence ``She wrote me a letter''.}
\label{fig:tree-morphing-example}
\end{figure}

\citet{sahin-emnlp18} introduce two operations to augment a dataset for low-resource POS tagging. Their method assumes access to a dependency tree, but they do not test it for dependency parsing, which we do here for the first time. The first operation, \textit{cropping}, removes some parts of a sentence to create a smaller or simpler, meaningful sentence. The second operation, \textit{rotation}, keeps all the words in the sentence but re-orders subtrees attached to the \textit{root} verb, in particular those attached by \textsc{nsubj} (nominal subject), \textsc{obj} (direct object), \textsc{iobj} (indirect object), or \textsc{obl} (oblique nominal) dependencies. Figure \ref{fig:tree-morphing-example} illustrates both operations. 

It is important to note that while both operations change the set of words or the word order, they do not change the dependencies. The sentences themselves may be awkward or ill-formed, but the corresponding analyses are still likely to be correct, and thus beneficial for learning. This is because they provide the model with more examples of variations in argument structure (cropping) and in constituent order (rotation), which may benefit languages with flexible word order and rich morphology. Some of our low-resource languages have these properties---while North Sámi has a fixed word order (SVO), Galician and Kazakh have relatively free word order. All three languages use case marking on nouns, so word order may not be as important for correct attachment. 

Both rotation and cropping can produce many trees. We use the default parameters given in \cite{sahin-emnlp18}.

\subsection{Data augmentation by nonce sentence generation (Nonce)}

Our next data augmentation method is adapted from \newcite{gulordava-naacl18}. The main idea is to create \textit{nonce} sentences by replacing some of the words which have the same syntactic annotations. For each training sentence, we replace each content word---nouns, verbs, or adjective---with an alternative word having the same universal POS, morphological features, and dependency label.\footnote{The dependency label constraint is new to this paper.} Specifically, for each content word, we first stochastically choose whether to replace it; then, if we have chosen to replace it, we uniformly sample the replacement word type meeting the corresponding constraints. For instance, given a sentence ``\textit{He borrowed a book from the library.}'', we can generate the following sentences:
\eenumsentence{
    \item\label{ex:nonce-1} \shortexnt{8} 
	{He & \underline{bought} & a & book & from & the & \underline{shop} & .}
	{}
	
	\item\label{ex:nonce-2} \shortexnt{8} 
	{He & \underline{wore} & a & \underline{umbrella} & from & the & library & .}
	{}
}
This generation method is only based on syntactic features (i.e., morphology and dependency labels), so it sometimes produces nonsensical sentences like \ref{ex:nonce-2}. But since we only replace words if they have the same morphological features and dependency label, this method preserves the original tree structures in the treebank. Following \citep{gulordava-naacl18}, we generate five nonce sentences for each original sentence.

\subsection{Cross-lingual training}
\label{sec:crossling}

When a source treebank is available, model transfer is a viable option. We perform model transfer by cross-lingual parser training: we first train on both source and target treebanks to produce a single model, and then fine tune the model only on the target treebank. In our preliminary experiments (Appendix \ref{appendix:fine-tuning}), we found that fine tuning on the target treebank was effective in all settings, so we use it in all applicable experiments reported in this paper.

\subsection{Transliteration}
\label{subsec:transliteration}

Two related languages might not share a writing system even when they belong to the same family. We evaluate whether a simple transliteration would be helpful for cross-lingual training in this case. In our study, the Turkish treebank is written in extended Latin while the Kazakh treebank is written in Cyrillic. This difference potentially makes model transfer less useful, and means we might not be able to leverage lexical similarities between the two languages. We pre-process both treebanks by transliterating them to the same ``pivot'' alphabet, basic Latin.\footnote{Another possible pivot is phonemes \citep{N16-1161}. We leave this as future work.} 

The mapping from Turkish is straightforward. Its alphabet consists of 29 letters, 23 of which are in basic Latin. The other six letters, `ç',`ğ', `ı', `ö', `ş', and `ü', add diacritics to basic Latin characters, facilitating different pronunciations.\footnote{https://www.omniglot.com/writing/turkish.htm} We map these to their basic Latin counterparts, e.g., `ç' to `c'.  For Kazakh, we use a simple dictionary created by a Kazakh computational linguist to map each Cyrillic letter to the basic Latin alphabet.\footnote{The mapping from Kazakh Cyrilic into basic Latin alphabet is provided in Appendix \ref{appendix:char-mapping}}.

\section{Experimental Setup}

\subsection{Dependency Parsing Model}
\label{sec:parsing-model}

We use the Uppsala parser, a transition-based neural dependency parser \citep{delhoneux17conll,delhoneux17arc,kiperwasser2016}. The parser uses an arc-hybrid transition system \citep{kuhlmann11}, extended with a static-dynamic oracle and \textsc{Swap} transition to allow non-projective dependency trees \citep{nivre09}. 

Let $w = w_0, \dots, w_{|w|}$ be an input sentence of length $|w|$ and let $w_0$ represent an artificial \textsc{Root} token. We create a vector representation for each input token $w_i$ by concatenating $(;)$ its word embedding, $\textbf{e}_w(w_i)$ and its character-based word embedding, $\textbf{e}_c(w_i)$:
\begin{align}
\label{eq:input-rep}
	\textbf{x}_i = [\textbf{e}_w(w_i);\textbf{e}_c(w_i)] 
\end{align}
Here, $\textbf{e}_c(w_i)$ is the output of a \textit{character-level} bidirectional LSTM (biLSTM) encoder run over the characters of $w_i$ \cite{ling2015}; this makes the model fully open-vocabulary, since it can produce representations for any character sequence. We then obtain a \emph{context-sensitive} encoding $\textbf{h}_i$ using a \textit{word-level} biLSTM encoder:
\begin{align}
\label{eq:context-biLSTM}
	\textbf{h}_i = [\textsc{LSTM}_f(\textbf{x}_{0:i});\textsc{LSTM}_b(\textbf{x}_{|w|:i})]
\end{align}
We then create a configuration by concatenating the encoding of a fixed number of words on the top of the stack and the beginning of the buffer. Given this configuration, we predict a transition and its arc label using a multi-layer perceptron (MLP). More details of the core parser can be found in \newcite{delhoneux17conll,delhoneux17arc}.

\subsection{Parameter sharing} 

To train cross-lingual models, we use the strategy of \citet{deLhoneux-emnlp18} for parameter sharing, which uses \textit{soft} sharing for word and character parameters, and \textit{hard} sharing for the MLP parameters. Soft parameter sharing uses a language embedding, which, in theory, learns what parameters to share between the two languages. Let $\textbf{c}_j$ be an embedding of character $c_j$ in a token $w_i$ from the treebank of language $k$, and let $\textbf{l}_k$ be the language embedding. For sharing on characters, we concatenate character and language embedding: $[\textbf{c}_j;\textbf{l}_k]$ for input to the character-level biLSTM. Similarly, for input to the word-level biLSTM, we concatenate the language embedding to the word embedding, modifying Eq. \ref{eq:input-rep} to
\begin{align}
\label{eq:word-emb}
    \textbf{x}_i = [\textbf{e}_w(w_i);\textbf{e}_c(w_i);\textbf{l}_k] 
\end{align}
We use the default hyperparameters of \citet{deLhoneux-emnlp18} in our experiments. We fine-tune each model by training it further only on the target treebank \citep{shi-etAl:ACL2016}. We use early stopping based on Label Attachment Score (LAS) on development set.

\subsection{Datasets}

We use Universal Dependencies (UD) treebanks version 2.2 \cite{UD2.2}. None of our target treebanks have a development set, so we generate new train/dev splits by 50:50 (Table \ref{tab:data}). Having large development sets allow us to perform better analysis for this study.

\begin{table}[t]
    \centering
    \small
    \begin{tabular}{llrrr}
        \toprule
        Language & Treebank ID & train & dev. & test \\
        \midrule
        Finnish & fi\_tdt & 14981 & 1875 & 1555 \\
        North Sámi & sme\_giella & 1128 & 1129 & 865 \\
        \midrule
        Portuguese & pt\_bosque & 8329 & 560 & 477 \\
        Galician & gl\_treegal & 300 & 300 & 400 \\
        \midrule
        Turkish & tr\_imst & 3685 & 975 & 975 \\
        Kazakh & kk\_ktb & 15 & 16 & 1047 \\
        \bottomrule
    \end{tabular}
    \caption{Train/dev split used for each treebank.}
    \label{tab:data}
\end{table}

\section{Parsing North Sámi}

\begin{table}[t]
    \centering
    \begin{tabular}{lrrr}
        \toprule
         & original & +Morph & +Nonce \\
        \midrule
        $\mathcal{T}_{100}$ & 1128 & 7636 & 4934 \\
        $\mathcal{T}_{50}$ & 564 & 3838 & 2700 \\
        $\mathcal{T}_{10}$ & 141 & 854 & 661 \\
        \bottomrule
    \end{tabular}
    \caption{Number of North Sámi training sentences.}
    \label{tab:ns-data}
\end{table}

North Sámi is our largest low-resource treebank, so we use it for a full evaluation and analysis of different strategies before testing on the other languages.
To understand the effect of target treebank size, we generate three datasets with different \textit{training} sizes: $\mathcal{T}_{10}$ (\mytilde10\%), $\mathcal{T}_{50}$ (\mytilde50\%), and $\mathcal{T}_{100}$ (100\%). Table \ref{tab:ns-data} reports the number of training sentences after we augment the data using methods described in Section \ref{sec:data-aug}. We apply \textsc{Morph} and \textsc{Nonce} separately to understand the effect of each method and to control the amount of noise in the augmented data.

\begin{table*}[t]
    \centering
    \begin{tabular}{lccc|cccc}
    \toprule
    \multicolumn{1}{c}{} & \multicolumn{3}{c}{\textsc{monolingual}} & \multicolumn{3}{c}{\textsc{cross-lingual}} \\
    \cmidrule{2-7}
    size & mono-base & +Morph & +Nonce & cross-base & +Morph & +Nonce  \\
    \toprule
    $\mathcal{T}_{100}$ & 53.3 & 56.0 (+3.3) & 56.3 (+3.0) & 61.3 (+8.0) & 60.9 (+7.6) & \textbf{61.7 (+8.4)} \\
    $\mathcal{T}_{50}$ & 42.5 & 46.6 (+4.1) & 46.5 (+4.0) & 52.0 (+9.5) & 51.7 (+9.2) & \textbf{52.0 (+9.5)} \\
    $\mathcal{T}_{10}$ & 18.5 & 27.1 (+8.6) & 27.8 (+9.3) & 34.7 (+16.2) & \textbf{37.3 (+18.8)} & 35.4 (+16.9) \\
    \bottomrule
    \end{tabular}
    \caption{LAS results on North Sámi development data. \textit{mono-base} and \textit{cross-base} are models without data augmentation. \% improvements over \textit{mono-base} shown in parentheses.}
    \label{tab:ns-results}
\end{table*}

We employ two baselines: a monolingual model (\textsection\ref{sec:parsing-model}) and a cross-lingual model (\textsection\ref{sec:crossling}), both \textit{without} data augmentation. The monolingual model acts as a simple baseline, to resemble a situation when the target treebank does not have any source treebank (i.e., no available treebanks from related languages). The cross-lingual model serves as a strong baseline, simulating a case when there is a source treebank. We compare both baselines to models trained with \textsc{Morph} and \textsc{Nonce} augmentation methods. Table \ref{tab:ns-results} reports our results, and we review our motivating scenarios below.

\paragraph{Scenario 1: we only have a very small target treebank.}
\noindent In the monolingual experiments, we observe that both dependency tree morphing (\textsc{Morph}) and nonce sentence generation (\textsc{Nonce}) improve performance, indicating the strong benefits of data augmentation when there are no other resources available except the target treebank itself. In particular, when the number of training data is the lowest ($\mathcal{T}_{10})$, data augmentations improves performance up to 9.3\% LAS.

\paragraph{Scenario 2: a source treebank is available.} We see that the cross-lingual training (cross-base) performs better than monolingual models even with augmentation. For the $\mathcal{T}_{10}$ setting, cross-base achieves almost twice as much as the monolingual baseline (mono-base). The benefits of data augmentation are less evident in the cross-lingual setting, but in the $\mathcal{T}_{10}$ scenario, data augmentation still clearly helps. Overall, cross-lingual combined with data augmentation yields the best result.

\subsection{What is learned from Finnish?}

Why do cross-lingual training and data augmentation help? To put this question in context, we first consider their relationship. Finnish and North Sámi are mutually unintelligible, but they are typologically similar: of the 49 (mostly syntactic) linguistic features annotated for North Sámi in the Word Atlas of Languages \citep[WALS;][]{wals}, Finnish shares the same values for 42 of them.\footnote{There are 192 linguistic features in WALS, but only 49 are defined for North Sámi. These features are mostly syntactic, annotated within different areas such as morphology, phonology, nominal and verbal categories, and word order.} Despite this and their phylogenetic and geographical relatedness, they share very little vocabulary: only 6.5\% of North Sámi tokens appear in Finnish data, and these words are either proper nouns or closed class words such as pronouns or conjunctions. However, both languages do share many character-trigrams (72.5\%, token-level), especially on suffixes.

Now we turn to an analysis of the $\mathcal{T}_{10}$ data setting, where we see the largest gains for all methods.

\subsection{Analysis of data augmentation}

For dependency parsing, POS features are important because they can provide strong signals whether there exists dependency between two words in a given sentence. For example, \textit{subject} and \textit{object} dependencies often occur between a \textsc{noun} and a \textsc{verb}, as can be seen in Fig. \ref{fig:tm-orig-sent}. We investigate the extent to which data augmentation is useful for learning POS features, using diagnostic classifiers \citep{Veldhoen2016DiagnosticCR,Adi2016FinegrainedAO,shi-etAl:ACL2016} to probe our model representations. Our central question is: do the models learn useful representations of POS, despite having no direct access to it? And if so, is this helped by data augmentation?

After training each model, we freeze the parameters and generate \textit{context-dependent} representations (i.e., the output of \textit{word-level} biLSTM, $\textbf{h}_i$ in Eq. \ref{eq:context-biLSTM}), for the training and development data. We then train a feed-forward neural network classifier to predict the POS tag of each word, using only the representation as input. To filter out the effect of cross-lingual training, we only analyze representations trained using the \textit{monolingual} models. Our training and development data consists of 6321 and 7710 tokens, respectively. The percentage of OOV tokens is 40.5\%.

\begin{table}[t]
    \centering
    \small
    \begin{tabular}{lrrrr}
    \toprule
    \multirow{2}{*}{POS} & \multirow{2}{*}{\%dev} & \multirow{2}{*}{baseline} & \multicolumn{2}{c}{\%diff. with} \\
    \cmidrule{4-5}
    & & & +Morph & +Nonce \\
    \toprule
    \textsc{intj} & 0.1 & 0.0 & 20.0 & 20.0 \\
    \textsc{part} & 1.5 & 70.1 & 7.7 & 0.8 \\
    \textsc{num} & 1.9 & 19.2 & 15.1 & -4.1 \\
    \textsc{adp} & 1.9 & 15.7 & 24.5 & 19.7 \\
    \textsc{sconj} & 2.4 & 57.8 & 5.9 & 7.6 \\
    \textsc{aux} & 3.2 & 26.3 & 27.2 & -4.9 \\
    \textsc{cconj} & 3.4 & 91.3 & -0.8 & -4.2 \\
    \textsc{propn} & 4.7 & 5.9 & 5.9 & -5.9 \\
    \textsc{adj} & 6.5 & 12.7 & 3.8 & 0.2 \\
    \textsc{adv} & 9.0 & 42.9 & 11.8 & 11.5 \\
    \textsc{pron} & 13.4 & 63.2 & 5.4 & -2.7 \\
    \textsc{verb} & 25.7 & 72.4 & -6.2 & -4.5 \\
    \textsc{noun} & 26.4 & 67.0 & 8.6 & 13.2 \\
    \bottomrule
    \end{tabular}
    \caption{Results for the monolingual POS predictions, ordered by the frequency of each tag in the dev split (\%dev). \%diff shows the difference between each augmentation method and monolingual models.}
    \label{tab:augmentation-analysis}
\end{table}

\begin{table*}[t]
    \centering
    \small
    \begin{tabular}{lll}
    \toprule
     & \multicolumn{2}{c}{Top nearest Finnish words} \\
    \cmidrule{2-3}
    North Sámi & char-level & word-level \\
    \midrule
     \textit{borrat} (\textsc{verb}; eat) &  \textit{herrat} (\textsc{noun}; gentleman) & \textit{käydä} (\textsc{verb}; go) \\
        & \textit{kerrat} (\textsc{noun}; time) & \textit{otan} (\textsc{verb}; take) \\
        & \textit{naurat} (\textsc{verb}; laugh) & \textit{sain} (\textsc{verb}; get) \\
    \midrule
    \textit{veahki} (\textsc{noun}; help) & \textit{nuuhki} (\textsc{verb}; sniff) & \textit{tyhjäksi} (\textsc{adj}; empty) \\
        & \textit{väki} (\textsc{noun}; power) & \textit{johonki} (\textsc{pron}; something) \\
        & \textit{avarsi} (\textsc{verb}; expand) & \textit{lähtökohdaksi} (\textsc{noun}; basis) \\
    \midrule
     \textit{divrras} (\textsc{adj}; expensive) & \textit{harras} (\textsc{adj}; devout) & \textit{välttämätöntä} (\textsc{adj}; essential) \\
        & \textit{reipas} (\textsc{adj}; brave) & \textit{mahdollista} (\textsc{adj}; possible) \\
        & \textit{sarjaporras} (\textsc{noun}; series) & \textit{kilpailukykyisempi} (\textsc{adj}; competitive)\\
    \midrule
    \end{tabular}
    \caption{Most similar Finnish words for each North Sámi word based on cosine similarity.}
    \label{tab:cosine-similarity-results}
\end{table*}

Table \ref{tab:augmentation-analysis} reports the POS prediction accuracy. We observe that representations generated with monolingual \textsc{Morph} seem to learn better POS, for most of the tags. On the other hand, representations generated with monolingual \textsc{Nonce} sometimes produce lower accuracy on some tags; only on nouns the accuracy is better than monolingual \textsc{Morph}. We hypothesize that this is because \textsc{Nonce} sometimes generates meaningless sentences which confuse the model. In parsing this effect is less apparent, mainly because monolingual \textsc{Nonce} has the poorest POS representation for infrequent tags (\%dev), and better representation of nouns.

\subsection{Effects of cross-lingual training}

Next, we analyze the effect of cross-lingual training by comparing the monolingual baseline to the cross-lingual model with \textsc{Morph}.

\paragraph{Cross-lingual representations.} 

The fact that cross-lingual model improves parsing performance is interesting, since Finnish and North Sámi have so little common vocabulary. What linguistic knowledge is transferred through cross-lingual training? We analyze whether words with the same POS category from the source and target treebanks have similar representations. To do this, we analyze the \textit{head predictions}, and collect North Sámi tokens for which only the cross-lingual model correctly predicts the headword.\footnote{Another possible way is to look at the label predictions. But since the monolingual baseline LAS is very low,  we focus on the unlabeled attachment prediction since it is more accurate.} For these words, we compare token-level representations of North Sámi \emph{development} data to Finnish \emph{training} data. 

We ask the following questions: Given the representation of a North Sámi word, what is the Finnish word with the most similar representation? Do they share the same POS category? Information other than POS may very well be captured, but we expect that the representations will reflect similar POS since POS is highly revelant to parsing. We use \textit{cosine distance} to measure similarity. 

We look at four categories for which cross-lingual training substantial improves results on the development set: adjectives, nouns, pronouns, and verbs. We analyze representations generated by two layers of the model in \textsection\ref{sec:parsing-model}: (1) the output of character-level biLSTM (char-level), $\textbf{e}_c(w_i)$ and (2) the output of word-level biLSTM (word-level), i.e., $\textbf{h}_i$ in Eq. \ref{eq:context-biLSTM}. 

Table \ref{tab:cosine-similarity-results} shows examples of top three closest Finnish training words for a given North Sámi word. We observe that character-level representation focuses on orthographic similarity of suffixes, rather than POS. On the word-level representations, we find more cases when the top closest Finnish words have the same POS with the North Sámi word. In fact, when we compare the most similar Finnish word (Table \ref{tab:top-1-cosine-sim}) quantitatively, we find that the word-level representations of North Sámi are often similar to Finnish word with the same POS; the same trend does not hold for character-level representations. Since very few word tokens are shared, this suggests that improvements in cross-lingual training might simply be due to syntactic (i.e. word order) similarities between the two languages, captured in the dynamics of the biLSTM encoder---despite the fact that it knows very little about the North Sámi tokens themselves. The word-level representation has advantage over the char-level representation in the way that it has access to contextual information like word order, and it has knowledge about the other words in the sentence.

\begin{table}[t]
    \centering
    \begin{tabular}{lcc}
    \toprule
    POS & char-level (\%) & word-level (\%)\\
    \toprule
    \textsc{adj} & 12.1 & 37.1 \\
    \textsc{noun} & 55.8 & 63.5 \\
    \textsc{pron} & 12.9 & 68.0 \\
    \textsc{verb} & 34.2 & 69.0 \\
    \bottomrule
    \end{tabular}
    \caption{\# of North Sámi tokens for which the most similar Finnish word has the same POS.}
    \label{tab:top-1-cosine-sim}
\end{table}

\begin{figure}[t]
    \includegraphics[width=\linewidth]{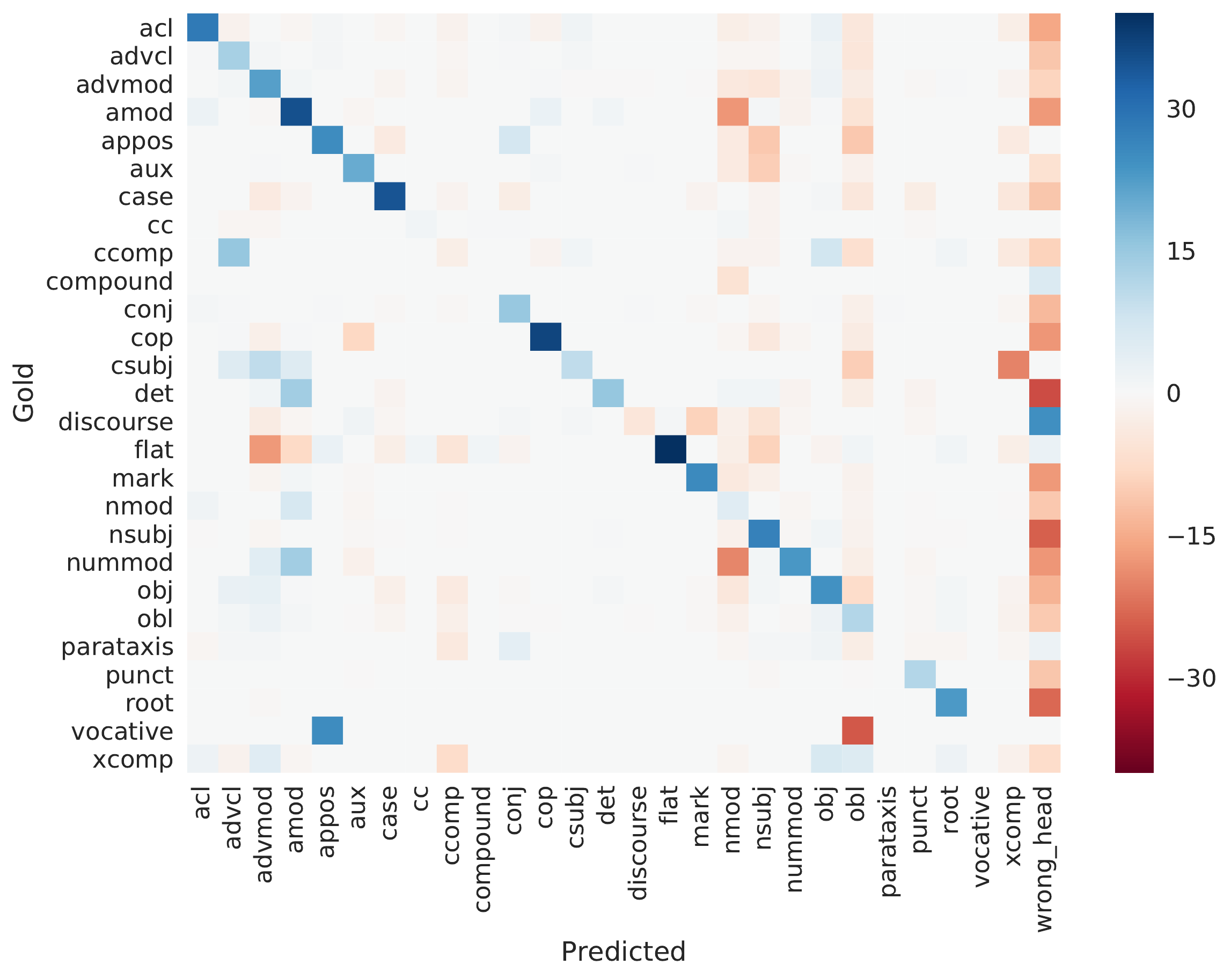}
    \caption{Differences between cross-lingual vs. monolingual confusion matrices. The last column represents cases of \textit{incorrect} heads and the other columns represent cases for \textit{correct} heads, i.e., each row summing to 100\%. Blue cells show higher cross-lingual values and red cells show higher monolingual values.}
    \label{fig:conf-matrix-deplbl}
\end{figure}

\paragraph{Head and label prediction.}

Lastly, we analyze the parsing performance of the monolingual compared to the cross-lingual models. Looking at the produced parse trees, one striking difference is that monolingual model sometimes predicts a ``rootless" tree. That is, it fails to assign a head of any word with index `0' and label the dependency with a \textit{root} label. In cases where the monolingual model predicts wrong parses and the cross-lingual model predicts the correct ones, we find that the ``rootless" trees are predicted more than 50\% of the time.\footnote{The parsing model enforces the constraint that every tree should have a head, i.e., an arc pointing from a dummy root to a node in the tree. It does not, however, enforce that this arc be labeled \textit{root}---the model must learn the labeling.} Meanwhile, the cross-lingual model learns to assign a word with head index `0’, although sometimes it is the incorrect word (e.g., it is the second word, but the parser predicts the fifth word). This pattern suggests that more training examples at least helps the model to learn structural properties of a well-formed tree.

The ability of a parser to predict labels is contingent on its ability to predict heads, so we focus our analysis on two  cases. How do monolingual and cross-lingual head prediction compare? And if both models predict the correct head, how do they compare on label prediction?

Figure \ref{fig:conf-matrix-deplbl} shows the \emph{difference} between two confusion matrices: one for cross-lingual and one for monolingual models. The last column shows cases of \textit{incorrect} heads and the other columns show label predictions when the heads are \textit{correct}, i.e., each row summing to 100\%. Here, blue cells highlight confusions that are more common for the cross-lingual model, while red cells highlight those more common for the monolingual model. For head prediction (last column), we observe that monolingual model makes higher errors especially for nominals and modifier words. In cases when both both models predict the correct heads, we observe that cross-lingual training gives further improvements in predicting most of the labels. In particular, regarding the ``rootless" trees discussed before, we see evidence that cross-lingual training helps in predicting the correct root index, and the correct \textit{root} label.

\begin{table}[t]
    \centering
    \small
    \begin{tabular}{lccc}
    \toprule
     & & \multicolumn{2}{c}{\textsc{cross-lingual}} \\
    \cmidrule{3-4}
    Language & zero-shot & +fastText & +Morph \\
    \midrule
    Galician & 51.9 & \textbf{72.8} & 71.0 \\
    Kazakh & 12.5 & 27.7 & \textbf{28.4} \\
    Kazakh (translit.) & 21.2 & 31.1 & \textbf{36.7} \\
    \bottomrule
    \end{tabular}
    \caption{LAS results on \textbf{development sets}. \textit{zero-shot} denotes results where we predict using model trained only on the source treebank.}
    \label{tab:lowres-results}
\end{table}

\section{Parsing truly low-resource languages}

\begin{table*}[t]
    \centering
    \begin{tabular}{lc|c|cc|c}
    \toprule
     \multicolumn{3}{c}{} & \multicolumn{2}{c}{\textsc{cross-lingual}} & \\
    \cmidrule{4-5}
    & baseline & best system & +fastText & +Morph & rank \\
    \midrule
    Galician & 66.16 & 74.25 & \textbf{70.46} & 69.21 & 10/27 \\
    Kazakh (translit.) & 24.21 & 31.93 & 25.28 & \textbf{28.23} & 2/27 \\
    \bottomrule
    \end{tabular}
    \caption{Comparison to CoNLL 2018 UD Shared Task on \textbf{test sets}. \textit{best system} is the state-of-the-art model for each treebank: UDPipe-Future \citep{udpipeUD18} for Galician and Uppsala \citep{uppsalaUD18} for Kazakh. \textit{rank} shows our best model position in the shared task ranking for each treebank.}
    \label{tab:conll-results}
\end{table*}

Now we turn to two truly low-resource treebanks: Galician and Kazakh. These treebanks are most analogous to the North Sami $\mathcal{T}_{10}$ setting and therefore we apply the best approach, cross-lingual training with \textsc{Morph} augmentation. Table \ref{tab:data} provides the statistics of the augmented data. For Galician, we use the Portuguese treebank as source while for Kazakh we use Turkish. Portuguese and Galician have high vocabulary overlap; 62.9\% of Galician tokens appear in Portuguese data, while for Turkish and Kazakh they do not share vocabulary since they use different writing systems. However, after transliterating them into the same basic Latin alphabet, we observe that 9.5\% of Kazakh tokens appear in the Turkish data. Both language pairs also share many (token-level) character trigrams: 96.0\% for Galician-Portuguese and 66.3\% for transliterated Kazakh-Turkish.

To compare our best approach, we create two baselines: (1) a pre-trained parsing model of the source treebank (zero-shot learning), and (2) a cross-lingual model initialized with \textit{monolingual} pre-trained word embeddings. The first serves as a weak baseline, in a case where training on the target treebank is not possible (e.g., Kazakh only has 15 sentences for training). The latter serves as a strong baseline, in a case when we have access to pre-trained word embeddings, for the source and/or the target languages. 

We treat a pre-trained word embedding as an external embedding, and concatenate it with the other representations, i.e., modifying Eq. \ref{eq:word-emb} to $\textbf{x}_i = [\textbf{e}_w(w_i);\textbf{e}_p(w_i);\textbf{e}_c(w_i);\textbf{l}_k]$, 
where $\textbf{e}_p(w_i)$ represents a pre-trained word embedding of $w_i$, which we update during training. We use the pre-trained monolingual fastText embeddings \cite{fastText}.\footnote{\raggedright The embeddings are available at https://fasttext.cc/docs/en/pretrained-vectors.html.} We concatenate the source and target pre-trained word embeddings.\footnote{If a word occurs in both source and target, we use the word embedding of the source language.} For our experiments with transliteration (\textsection\ref{subsec:transliteration}), we transliterate the entries of both the source and the target pre-trained word embeddings.

\subsection{Experimental results} 

Table \ref{tab:lowres-results} reports the LAS performance on the development sets. \textsc{Morph} augmentation improves performance over the zero-shot baseline and achieves comparable or better LAS with a cross-lingual model trained with pre-trained word embeddings. 

Next, we look at the effects of transliteration (see Kazakh vs Kazakh (translit.) in Table \ref{tab:lowres-results}). In the zero-shot experiments, simply mapping both Turkish and Kazakh characters to the Latin alphabet improves accuracy from 12.5 to 21.2 LAS. Cross-lingual training with \textsc{Morph} further improves performance to 36.7 LAS.

\subsection{Comparison with CoNLL 2018} 

To see how our best approach (i.e., cross-lingual model with \textsc{Morph} augmentation) compares with the current state-of-the-art models, we compare it to the recent results from CoNLL 2018 shared task. Training state-of-the-art models may require lots of engineering and data resources. Our goal, however, is not to achieve the best performance, but rather to systematically investigate how far simple approaches can take us. We report performance of the following: (1) the shared task baseline model \citep[UDPipe v1.2;][]{udpipe:2017} and (2) the best system for each treebank, (3) our best approach, and (4) a cross-lingual model with fastText embeddings.

Table \ref{tab:conll-results} presents the overall comparison on the test sets. For each treebank, we apply the same sentence segmentation and tokenization used by each best system.\footnote{UD shared task only provides unsegmented (i.e., sentence-level and token-level) raw test data. However, participants were allowed to use predicted segmentation and tokenization provided by the baseline UDPipe model.} We see that our approach outperforms the baseline models on both languages. For Kazakh, our model (with transliteration) achieves a competitive LAS (28.23), which would be the second position in the shared task ranking. As comparison, the best system for Kazakh \citep{uppsalaUD18} trained a multi-treebank model with four source treebanks, while we only use one source treebank. Their system use predicted POS as input, while ours depends solely on words and characters. The use of more treebanks and predicted POS is beyond the scope of our paper, but it is interesting that our approach can achieve the second best with such minimal resources. For Galician, our best approach outperforms baseline by 8.09 LAS points. Note that, Galician treebank does not come with training data. We use 50:50 train/dev split, while other teams might use higher split for training (for example, the best system \citep{udpipeUD18} uses 90:10 train/dev split). Since we treat Galician as our test data, we did not tune on the proportion for training data, but we guess that this is the main reason why our system achieve rank 10 out of 27.

Compared to cross-lingual models with fastText embeddings (fastText vs. \textsc{Morph}), we observe that our approach achieve better or comparable performance, showing its potential when there is not enough monolingual data available for training word embeddings.

\section{Conclusions}

In this paper, we investigated various low-resource parsing scenarios. We demonstrate that in the extremely low-resource setting, data augmentation improves parsing performance both in monolingual and cross-lingual settings. We also show that transfer learning is possible with \textit{lexicalized} parsers. In addition, we show that transfer learning between two languages with different writing systems is possible, and future work should consider transliteration for other language pairs.

While we have not exhausted all the possible techniques (e.g., use of external resources \citep{TACL922,K18-2019}, predicted POS \citep{ammar-tacl16}, multiple source treebanks \citep{L18-1352,stymne-emnlp18}, among others), we show that simple methods which leverage the linguistic annotations in the treebank can improve low-resource parsing. Future work might explore different augmentation methods, such as the use of \textit{synthetic} source treebanks \citep{D18-1163} or contextualized language model \citep{ELMO,ULMFit,devlin2018bert} for scoring the augmented data (e.g., using perplexity).

Finally, while the techniques presented in this paper might be applicable to other low-resource languages, we want to also highlight the importance of understanding the characteristics of languages being studied. For example, we showed that although North Sami and Finnish do not share vocabulary, cross-lingual training is still helpful because they share similar syntactic structures. Different language pairs might benefit from other types of similarity (e.g., morphological) and investigating this would be another interesting future work for low-resource dependency parsing.

\section*{Acknowledgments}

Clara Vania is supported by the Indonesian Endowment Fund for Education (LPDP), the Centre for Doctoral Training in Data Science, funded by the UK EPSRC (grant EP/L016427/1), and the University of Edinburgh. Anders S{\o}gaard is supported by a Google Focused Research Award. We thank Aibek Makazhanov for helping with Kazakh transliteration, and Miryam de Lhoneux for parser implementation. We also thank Andreas Grivas, Maria Corkery, Ida Szubert, Gozde Gul Sahin, Sameer Bansal, Marco Damonte, Naomi Saphra, Nikolay Bogoychev, and anonymous reviewers for helpful discussion of this work and comments on previous drafts of the paper. 

\bibliography{emnlp-ijcnlp-2019}
\bibliographystyle{acl_natbib_nourl}

\clearpage
\appendix

\section{Effects of Fine-Tuning for Cross-Lingual Training}
\label{appendix:fine-tuning}

For our cross-lingual experiments in Section \ref{sec:crossling}, we observe that fine-tuning on the target treebank always improves parsing performance. Table \ref{tab:fine-tuning-exp} reports LAS for cross-lingual models with and without fine-tuning.

\begin{table}[ht]
    \centering
    \small
    \begin{tabular}{l|ccc}
    \toprule
    size & cross-base & +Morph & +Nonce \\
    \midrule
    $\mathcal{T}_{100}$ & 57.9 (+4.6) & 59.5 (+6.2) & 59.3 (+6.0) \\
    $\mathcal{T}_{50}$ & 48.3 (+5.8) & 49.8 (+7.3) & 50.1 (+7.6) \\
    $\mathcal{T}_{10}$ & 29.8 (+11.3) & 34.9 (+16.4) & 34.8 (+16.3) \\
    \midrule
     & \multicolumn{3}{c}{\textit{$\downarrow$ with fine tuning (FT) $\downarrow$}} \\
    \midrule
    $\mathcal{T}_{100}$ & 61.3 (+8.0) & 60.9 (+7.6) & \textbf{61.7 (+8.4)} \\
    $\mathcal{T}_{50}$ & 52.0 (+9.5) & 51.7 (+9.2) & \textbf{52.0 (+9.5)} \\
    $\mathcal{T}_{10}$ & 34.7 (+16.2) & \textbf{37.3 (+18.8)} & 35.4 (+16.9) \\
    \bottomrule
    \end{tabular}
    \caption{Effects of fine-tuning on North Sámi development data, measured in LAS. \textit{mono-base} and \textit{cross-base} are models without data augmentation. \% improvements over \textit{mono-base} shown in parentheses.}
    \label{tab:fine-tuning-exp}
\end{table}

\section{Cyrillic to Latin Alphabet mapping}
\label{appendix:char-mapping}

We use the following character mapping for Cyrillic to Latin Kazakh treebank transliteration. 

\begin{figure}[ht]
    \centering
    \includegraphics[width=.9\linewidth]{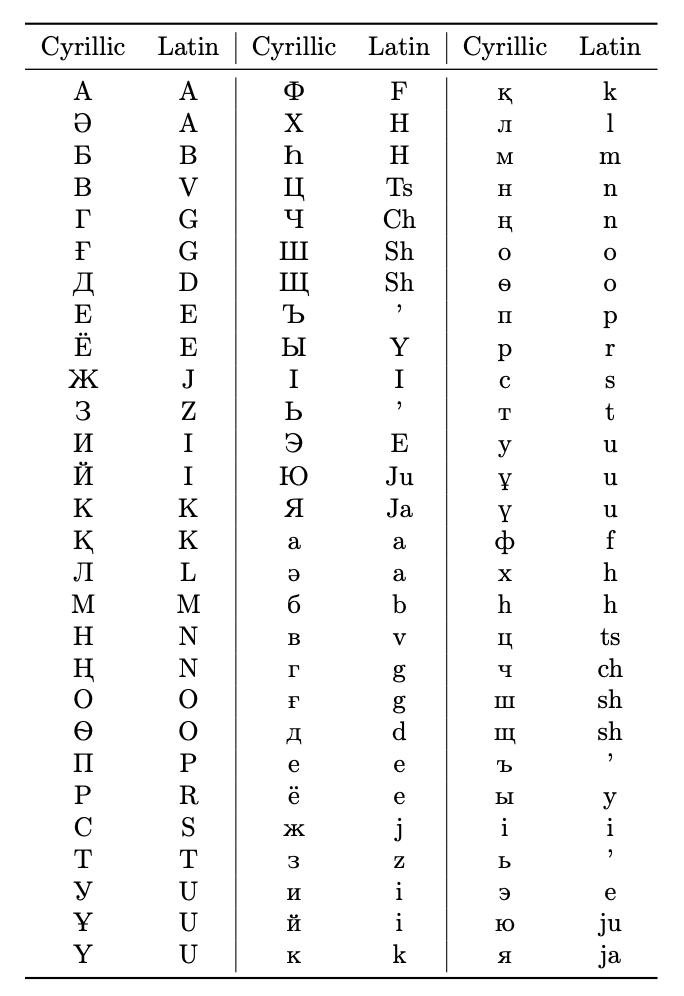}
    \caption{Cyrillic to Latin alphabet mapping.}
    \label{fig:alph-mapping}
\end{figure}

\end{document}